\pdfoutput=1

\documentclass[11pt]{article}

\usepackage{EMNLP2022}
\usepackage{times}
\usepackage{latexsym}
\usepackage{caption}
\usepackage{subcaption}
\usepackage[T1]{fontenc}

\usepackage[utf8]{inputenc}

\usepackage{microtype}
\usepackage{graphicx}

\usepackage{inconsolata}

\title{Template-based Recruitment Email Generation for Job Recommendation}

\author{Qiuchi Li \and Christina Lioma\\
  Department of Computer Science \\
  University of Copenhagen \\
  \{qiuchi.li, c.lioma\}@di.ku.dk
  }

\begin{document}
\maketitle
\begin{abstract}
Text generation has long been a popular research topic in NLP. However, the task of generating recruitment emails from recruiters to candidates in the job recommendation scenario has received little attention by the research community. This work aims at defining the topic of automatic email generation for job recommendation, identifying the challenges, and providing a baseline template-based solution for Danish jobs. Evaluation by human experts shows that our method is effective. We wrap up by discussing the future research directions for better solving this task.
\end{abstract}

\section{Introduction and Prior Work}

Recruitment email generation is a crucial step in the overall job recruitment process. When recruiters find suitable candidates for a job, they need to write emails to contact the candidates, explaining why they are fit for the job and inviting them to apply. The job market produces a huge number of job postings on a daily basis, which results in a significant amount of human labor required to write recruitment emails. The recruiters are in urgent need of an automatic approach to compose these emails, to reduce their efforts and increase productivity.

The core challenges facing the task are two-fold: \textbf{persuasiveness} and \textbf{personalization}. First, the emails should contain sufficient reasons to convince the candidate that their qualifications match the requirements of the position, and illustrate that the position meets the candidate's expectations. Second, personalized emails should let the job seekers feel that the recruiters pay special attention to them and in turn motivate them to apply. In an interview of recruiters of varying recruitment experience, recruiters believe that proper personalization on the recruitment emails could boost candidate positive response rate, and admit that the current emails, largely composed based on fixed templates, are not personalized and persuasive enough~\cite{bogers_exploration_2021}. 

Recruitment email generation can be seen as a task-oriented text generation problem. One needs to extract information from the input job description and candidate profile, and generate the recruitment email accordingly. A large body of research on emails focuses on the analysis side, typically automatic detection of phishing emails by extracting features~\cite{Basnet2008, Verma_2017_phish, yu_2009_phishcatch}. In the realm of email synthesis, email subject line generation~\cite{xue_2020_key,zhang_2019_email} has been studied. For generation of the main email body, attempts have been made 
on fake email generation for malicious purposes~\cite{das_2019_automated,baki_2017_scaling} based on a two-step pipeline~\cite{chen-2014-two,chen-2014-two-longtitle} for email synthesis. As far as we know, generating recruitment emails in  job recruitment is an unexplored task in literature, despite its  practical significance and challenging nature.

We therefore formally formulate the task of recruitment email generation and provide a baseline template-based system for it by extending~\citet{chen-2014-two}'s approach. In particular, we randomly generate an email from a library of different pre-defined components, and fill the motivational sentence with certain slots by combining matched skills and occupations extracted from the job and candidate in question. We conduct a user study to evaluate the quality of the generated text and examine if it can save recruiters' time in writing emails. The results show that recruiters under test are overall satisfied with the generated emails, and the time spent on writing emails for each candidate is significantly reduced.

The rest of the paper is organized as follows. We define the problem formally in Section 2. We run a pilot study to examine the simple end-to-end neural generation algorithm in Section 3 and elicit the need for a fine-grained synthesizing approach. Our template-based approach is described in Section 4, and the user evaluation is reported and discussed in Section 5. We conclude and point out future directions in Section 6.
\section{Task Definition and Notations}

Recruitment email generation calls for generating an email based on an input the job and a candidate . It requires us to extract the matched information from different job and candidate fields, and convert it to natural language expressions in the email. Typically, a job posting contains its title, company name and a textual description. A candidate profile has a headline, a list of keywords, a list of preferred job titles, the candidate's work experience and educational experience, and the candidate's resume text.

\begin{figure*}[ht]
\caption{Example of a real recruitment email for the given job and  candidate pair. The texts are translated from Danish to English.}
\label{fig:example}
\centering
\includegraphics[width=0.95\textwidth]{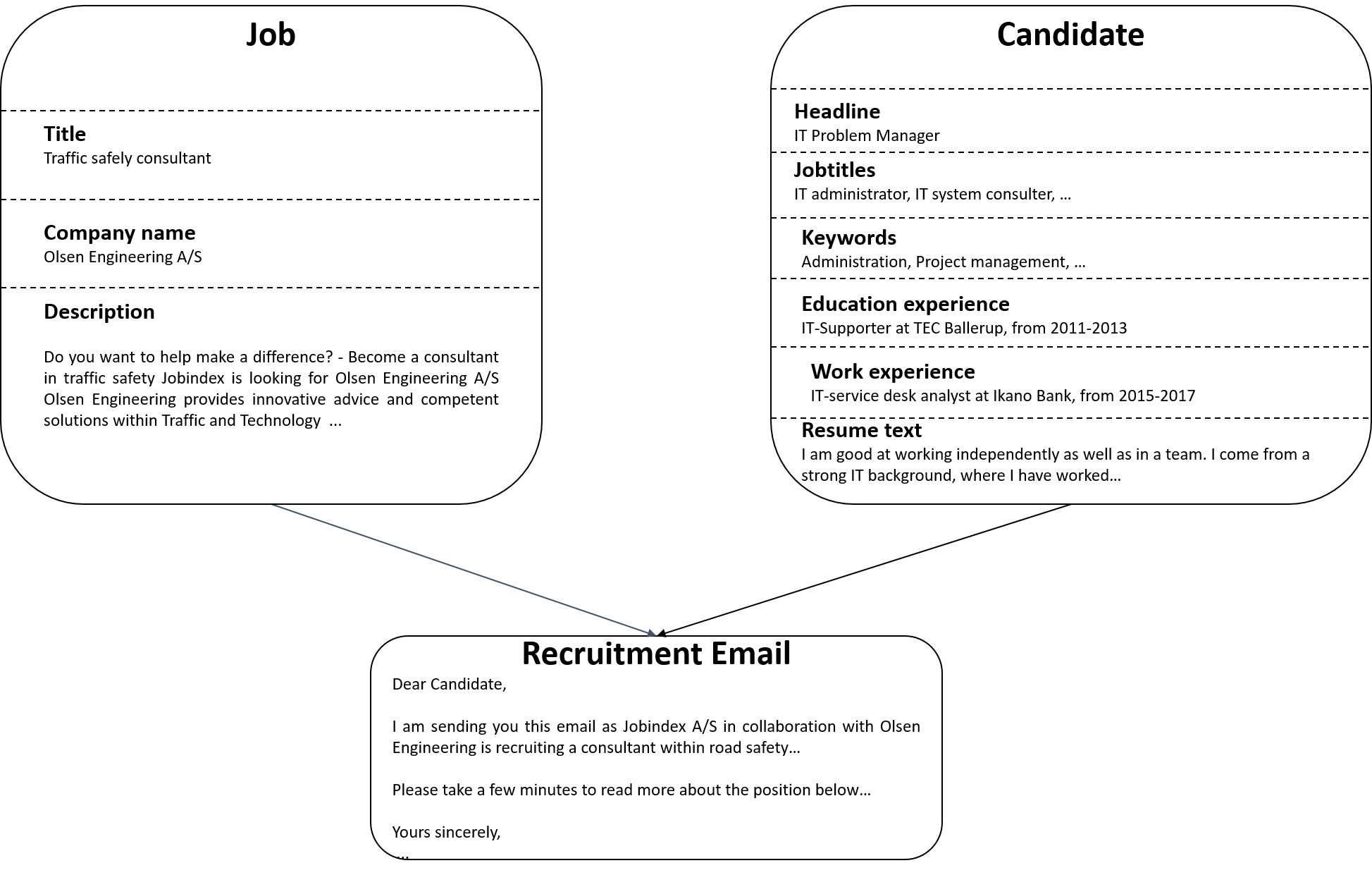}
\end{figure*}

The data we work with comes from Jobindex\footnote{https://www.jobindex.dk/}, one of Scandinavia’s largest job portals and recruitment agencies. In the Jobindex system, the above-mentioned fields are the main source of information for jobs and candidates. The majority of the jobs and candidate profiles are in Danish.  Please refer to Fig.~\ref{fig:example} for an example of a pair of job and candidate, and the real recruitment email written by a human recruiter. The personal information has been removed from the example and all texts are translated to English.
\section{Pilot Study: end-to-end neural generation}

The most intuitive way to handle the recruitment email generation task is to generate the whole email in a fully end-to-end fashion by concatenating the job and company text. We conduct a pilot study to examine whether this setting works in practice.

\textbf{Model}. We build a Transformer encoder-decoder, the state-of-the-art architecture for natural language generation, to support a sequence-to-sequence generation of recruitment email. The job and candidate texts are fed to the encoder side, and the recruitment email is generated in a token-by-token auto-regressive manner. The Transformer structure enables us to load the weights of the pre-trained Danish language model, Danish-BERT~\footnote{https://huggingface.co/Maltehb/danish-bert-botxo}.
~\citet{rothe-etal-2020-leveraging} conducted a comprehensive comparison of different strategies to use the pre-trained BERT weights for sequence-to-sequence generation, and found that the \textit{bert2bert} setting can yield robust performance across different text generation tasks. Therefore, we follow this setting and initialize the weights for both encoder and decoder with the Danish-BERT checkpoint. Please refer to~\citet{rothe-etal-2020-leveraging} for further details. We claim that this model may not be the state-of-the-art for Danish text generation, since Danish GPT is publicly available~\footnote{https://huggingface.co/flax-community/dansk-gpt-wiki}. However, it is a strong-performing model due to its Transformer architecture, and its generated text is representative of the state-of-the-art neural generation models.

\textbf{Data Preprocessing}. We focus on generating Danish emails from Danish jobs and candidate profiles. We mine (job, candidate, email) triplets from the Jobindex database. We filter out samples with non-Danish text, too short job descriptions and empty job titles. A total number of 317566 samples are obtained. Due to the limitation on input length by the model, we concatenate the summary of job description and the headline, job titles, education experience and work experience of candidate. Different fields are split by the periods. For recruitment emails, we only take their main bodies. We replace specific job title and company name with special tokens ``[job]'' and ``[cpy]'' respectively. 

\textbf{Training and Evaluation}. Our Transformer encoder-decoder has 12 layers and 12 attention heads and a maximum length of 512 tokens on both encoder and decoder sides. The embedding dimension and hidden dimension are set to 768 and 3072, respectively. We split all samples into training, validation and test set at a 7:2:1 ratio. We perform mini-batch learning to train the model with a batch size of 32. The average cross-entropy loss over all tokens at the decoder side is used as the loss function. We train the model on the training set for a maximum number of 5 epochs with Adam optimizer, and stop training when the validation loss stops dropping.

We use bilingual evaluation understudy (BLEU) as the quantitative metric for the generated message. The BLEU scores are calculated for individual translated segments, by comparing them with a set of good quality reference translations. The average scores over the whole test corpus are computed as an estimation of the overall quality. 

\textbf{Result}. We obtained a BLEU score of 23.2 on test samples for the generated texts, which is far from the SOTA neural text generation system for Danish (32.5 - 33.8)~\cite{fan-gardent-2020-multilingual}. This reveals that the generated emails have a limited overall quality. 

\begin{figure*}[ht]
\caption{Examples of generated emails by the end-to-end neural model. The generated motivational sentences are translated in English and colored in red. }
\label{fig:generated_examples}
\centering
\includegraphics[width=0.95\textwidth]{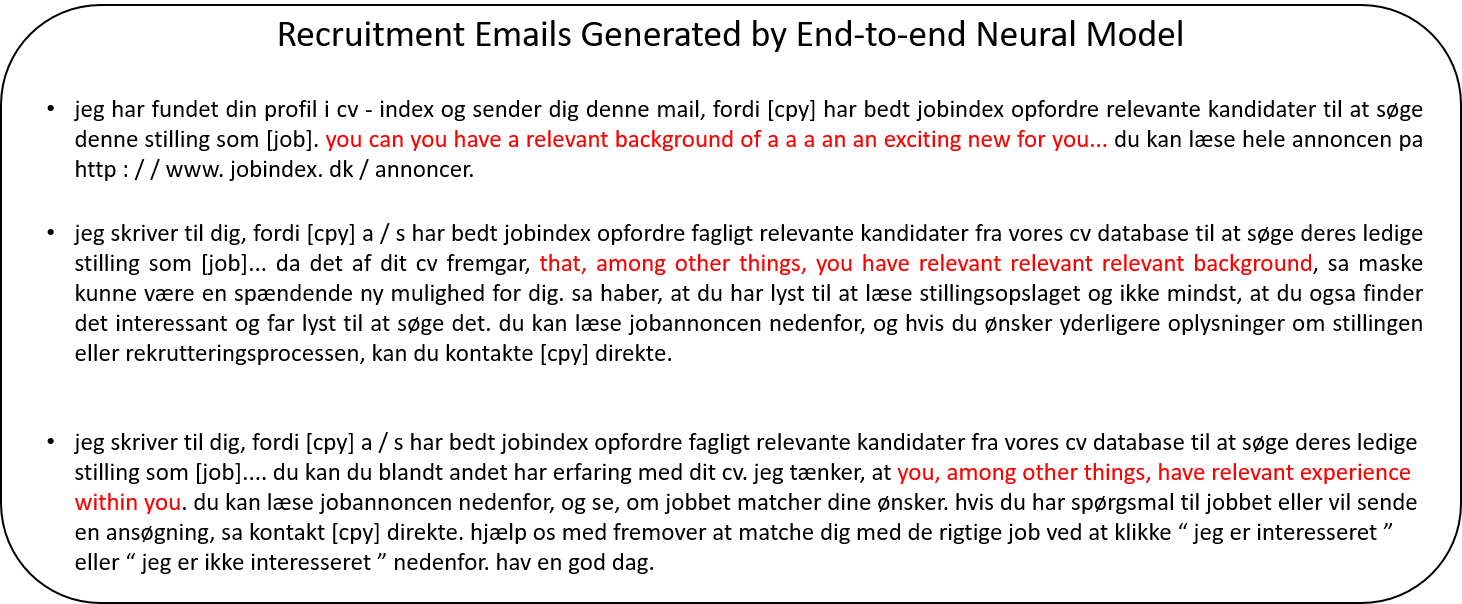}
\end{figure*}

Additionally, manual inspection shows that the auto-generated emails have low extent of personalization and persuasiveness. The generated motivational sentences usually do not contain specific reasons of match and are prone to grammatical errors. See Fig.~\ref{fig:generated_examples} for concrete examples. We posit that it is because the motivational sentences are usually located in the middle of email body, embedded in texts primarily generated from templates. Also, over 80\% of the motivational sentences are devoid of case-specific information such as the matched skills or occupations. Therefore, an end-to-end generation system is capable of learning these recurring template-based texts quite well, but poor at learning  case-specific motivational sentences.

\section{Proposed Methodology}

\begin{figure*}[h]
\caption{Diagram of the email generation system. The final generated messages are the main body of a real recruitment email, translated to English for this example.}
\label{fig:template-based-generation-pipeline}
\centering
\includegraphics[width=0.95\textwidth]{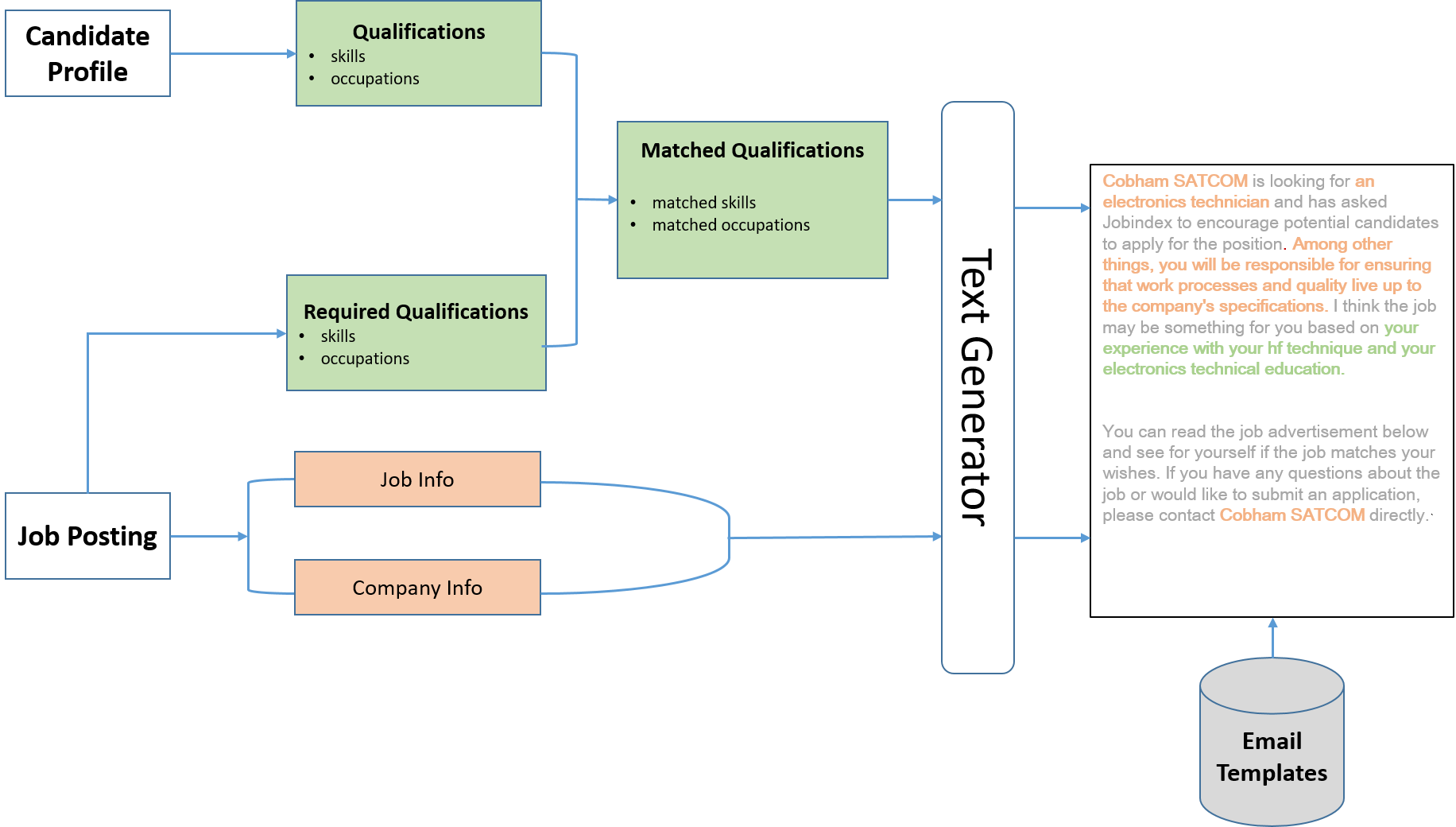}
\end{figure*}

The evaluation result of end-to-end neural generation indicates the need for a finer-grained generation system. Instead of generating the email as a whole, a special module should be developed to generate case-specific information, such as the motivational sentence, from the input job and candidate texts. We are inspired by the two-step email synthesis~\cite{chen-2014-two} to generate recruitment emails: email structure generation and slot filling. Specifically, we randomly generate an email template from a pre-constructed library of email components. The slots of the template indicate case-specific fields, and are then filled by extracting information from the matched job and candidate. The overall process of the system is shown in Fig.~\ref{fig:template-based-generation-pipeline}.

Potentially, our approach is superior to end-to-end generation in three aspects: 1) recruiters follow a similar process to write emails in a real-case scenario, so the generated emails are more likely to be accepted by and benefit recruiters; 2) the algorithm has a better control of the generated content, and ensures a certain extent of grammatical correctness and readability; 3) by explicitly composing case-specific information, the algorithm enhances personalization and persuasiveness of the generated recruitment emails.

In the following text, we introduce the implementation details of our template-based generation system.

\subsection{Template Parsing}

\textbf{Template data cleaning.} We get a total number of 553 raw templates from Jobindex system. We remove the non-Danish templates and templates for other purposes from the collection, and the remaining ones are classified into company-specific and general templates. Company-specific templates are provided by some companies to the recruiters with the request that these templates are strictly used when generating recruitment emails for these specific companies. General templates are used for all other cases. 
Eventually, 270 general templates and 81 specific templates remain in the database.\\

\noindent \textbf{Manual annotation.}  We define a list of email components (or fields) according to their functions based on manual inspection. They are mainly categorized into \textbf{functional}, \textbf{case-specific} and \textbf{auto-fill} fields. Functional fields are indicative of email structure, and do not entail case-specific information. Case-specific fields contain information specific to the matched pair of job and candidate, such as the job title, company name, motivational sentences expressing why the candidate is fit for the job, and so on. Some case-specific fields are automatically filled by the Jobindex system, and they are referred to as auto-fill fields instead. A total number of 54 email components are defined.

For each template in the database, we manually annotate the text by inserting HTML tags before and after a certain text segment to indicate its function in the email.  

\begin{figure*}[h]
\caption{The HTML tag, description and example of email components, extracted from the template database. }
\label{fig:email_components}
\centering
\includegraphics[width=0.95\textwidth]{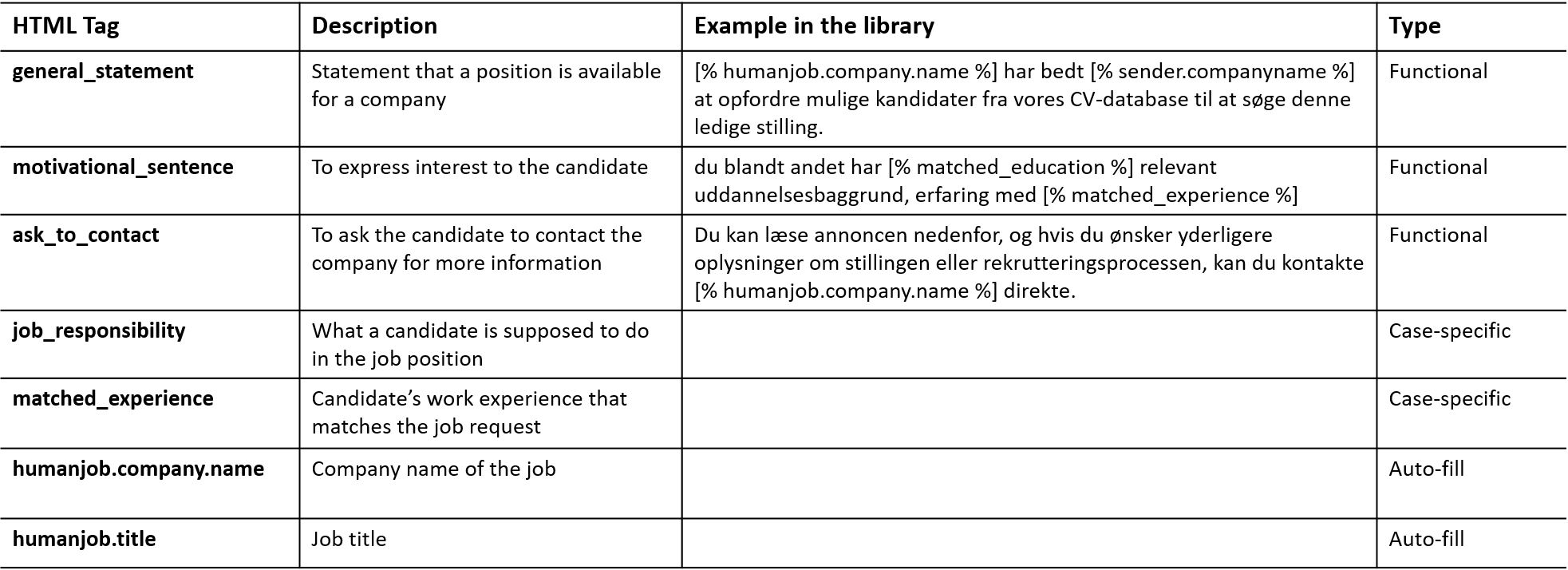}
\end{figure*}

\noindent \textbf{Parsing.} We parse the annotated templates to construct the email component library, essentially a dictionary of (component\_name, content list) pairs. The annotated templates are each parsed in a recursive fashion by an HTML parser. Starting from the root element of the whole template, the parser performs the same operations for each element: identify its child elements, process each the child element, append the processed child element content to the dictionary, and replace the child element with the marker ``\textbf{[\% f \%]}''. As such, the parser navigates all HTML tags in the template and adds their contents to the corresponding component content list. In the text content of each component, its child components are all masked with ``\textbf{[\% f \%]}'' tags as desired. The output of the algorithm is the top-level structure of the email templates, which are stored in the dictionary as values of ``\textbf{skeleton\_follower}'' and ``\textbf{skeleton\_non\_follower}'' for followers and for non-followers, respectively. Followers are candidates who follow a company in their profile, and non-followers are candidates who do not follow a company. After parsing all templates, we obtain a list of contents for each pre-defined email component. Examples of extracted email components are shown in Fig.~\ref{fig:email_components}. It is worth noting that nested structures universally appear among functional components, i.e. the textual content of one component may contain other components.\\

\subsection{Baseline Template-based Email Generation Algorithm}

We build an email generation system based on the constructed email component library. The algorithm randomly generates a template from the constructed library, based on whether the database contains the company name and whether the candidate is a follower of the company. Then, we extract the matched skills and occupations from the input job and candidate, and convert them to a coherent expression for the motivational sentence explaining why the candidate fits the job. \\

\noindent \textbf{Random template generation}.  Two issues should be accounted for when generating a proper template for a matched (job, candidate) pair. First, some companies prefer writing recruitment emails based on company-specific templates, so the system should always use a company template where possible. Second, followers should be contacted with slightly different emails to stress that the candidate follows the company.



The algorithm works as follows. First, we search for templates with the company name of the job posting. If company-specific templates can be found, then we randomly choose a template from the matched company templates. If it returns an empty list, then we randomly generate a template from the email component library. We start with a randomly selected follower skeleton or a non-follower skeleton according to whether the company is in the candidate's following list. Then, we navigate through all unfilled slots in an iterative fashion. For each slot, we randomly pick a text content from all its respective candidates in the library. Since the text expression may contain other unfilled slots, this process goes in an iterative fashion until no functional component tags appear in the templates.

\begin{figure}[ht]
\caption{A randomly generated template for non-followers.}
\label{fig:random_generate_template}
\centering
\includegraphics[width=0.5\textwidth]{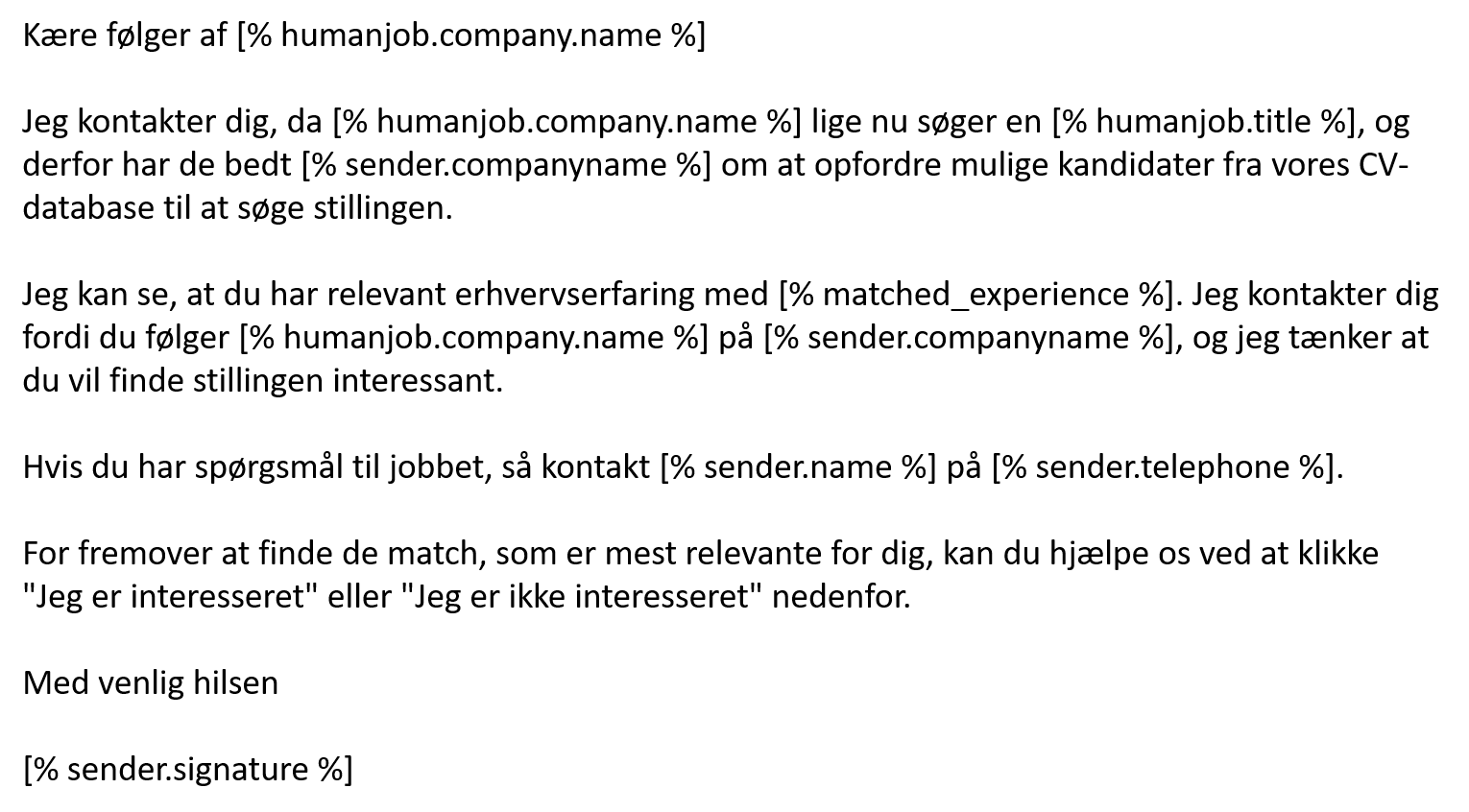}
\end{figure}

To this end, we have obtained a template for the recruitment email. An example can be seen in Fig.~\ref{fig:random_generate_template}. \\

\noindent \textbf{Template filling}. We compose the motivational sentence based on the matched qualifications of the candidate, including skills and occupations. For this purpose, we construct a dictionary of skills and occupations from the ESCO (European Skills, Competences, Qualifications and Occupations) database\footnote{https://esco.ec.europa.eu/}. ESCO has a 4-level category annotation for skills and occupations in different languages. We include both skills and occupations as dictionary keys in English and Danish. We also add the list of IT-related skills from Microsoft Skills library\footnote{https://learn.microsoft.com/en-us/} into the dictionary. A total number of 129474 skills and 46060 occupations are collected.

Based on the dictionary, we extract skills and occupations from the matched pair of job and candidate texts and obtain their intersections. We first use a built-in name entity recognition (NER) algorithm in the spaCy\footnote{https://spacy.io/} package to identify text spans that may contain skills and occupations, and check whether each is a skill or an occupation. The matched skills and occupations are obtained by taking the set intersections. We further split the language skills from the matched skills set. 

Simple rules are then applied to convert the matched skills, occupations and language skills to a coherent expression as explanations of why the candidate fits the job. The skills and occupations are respectively concatenated with the ``\{\}, \{\} og \{\}'' pattern, and conjunctive expressions are added to combine the expressions for skill and occupation match. It is further appended by the language skill match expression, if non-Danish skills are detected in both the job and candidate. The whole expression is inserted to the motivation sentence template with modifications of the pre-slot expressions to ensure coherence. When multiple slots appear in a motivational sentence template, they are inserted with matched skills and matched occupations, respectively. As a back-up, a general text expression will be randomly chosen from a list of contents in case of no matched qualifications.

\begin{figure}[ht]
\caption{An example of composed motivational sentence from matched skills, occupations and language skills. }
\label{fig:composed_message}
\centering
\includegraphics[width=0.47\textwidth]{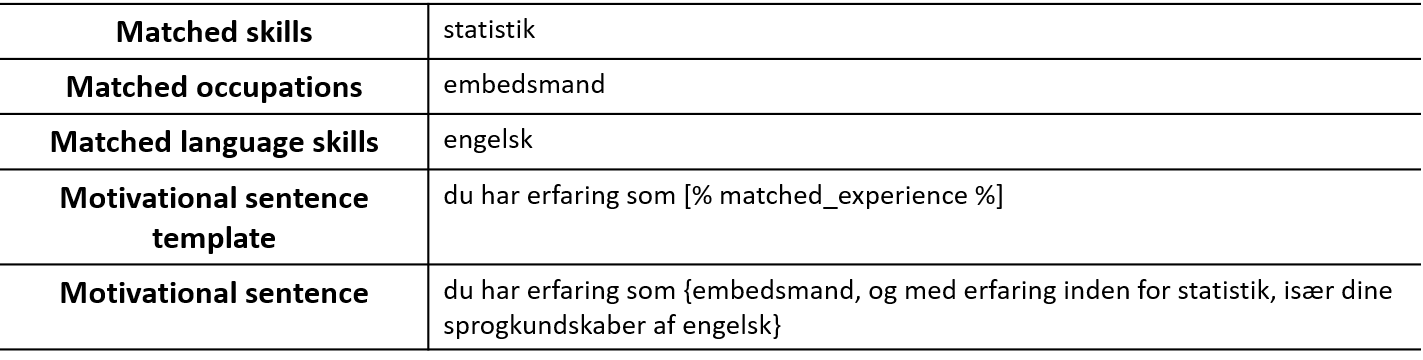}
\end{figure}

Please see Fig.~\ref{fig:composed_message} for a visual illustration of how the matched qualifications are converted to a motivational expression.

\noindent \textbf{Text post-processing}. We automatically post-process the generated email in the final step. Specifically, if multiple punctuations appear in a row, we keep only the last punctuation. We also make sure the first letter of a sentence is upper-cased. Finally, we correct the spacing errors between words in a sentence, between sentences, and between paragraphs. This gives the final output message produced by the template-based recruitment email generation algorithm.

\section{Evaluation}

\begin{figure*}
     \centering
     \begin{subfigure}[b]{0.47\textwidth}
         \centering
         \includegraphics[width=\textwidth]{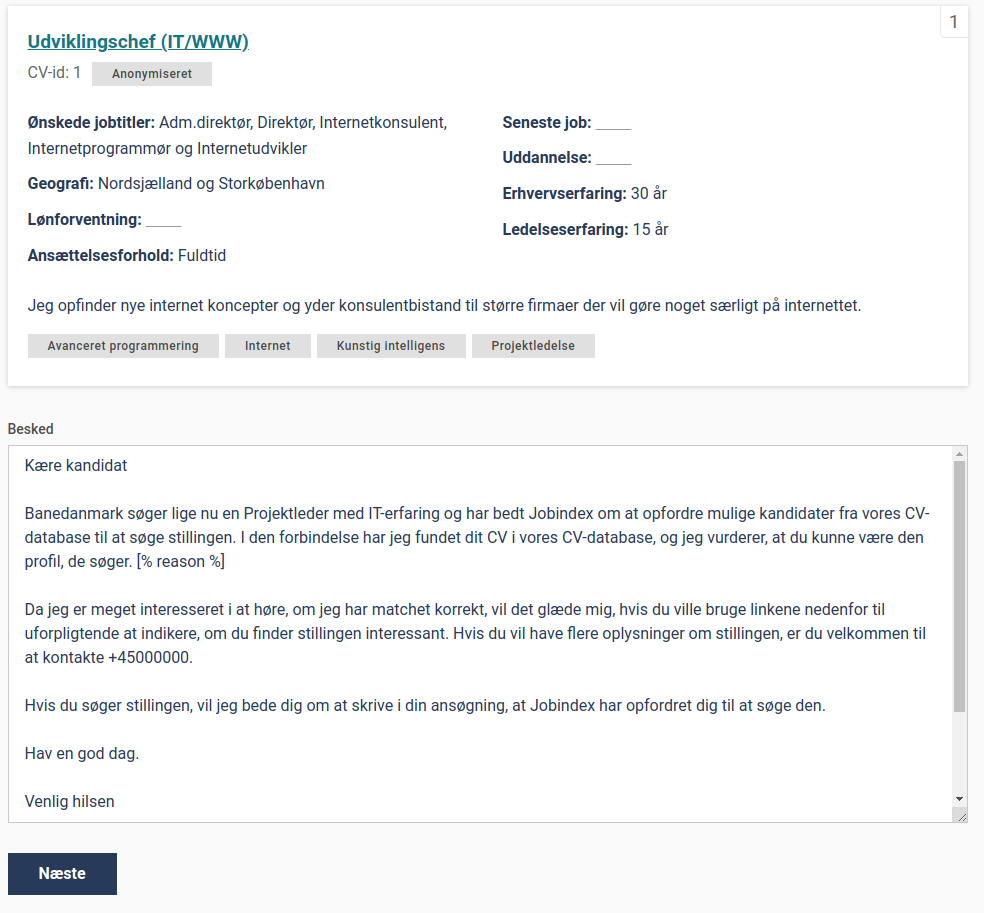}
         \caption{Unhelped case. }
         \label{fig:unhelped-after}
     \end{subfigure}
     \hfill
     \begin{subfigure}[b]{0.47\textwidth}
         \centering
         \includegraphics[width=\textwidth]{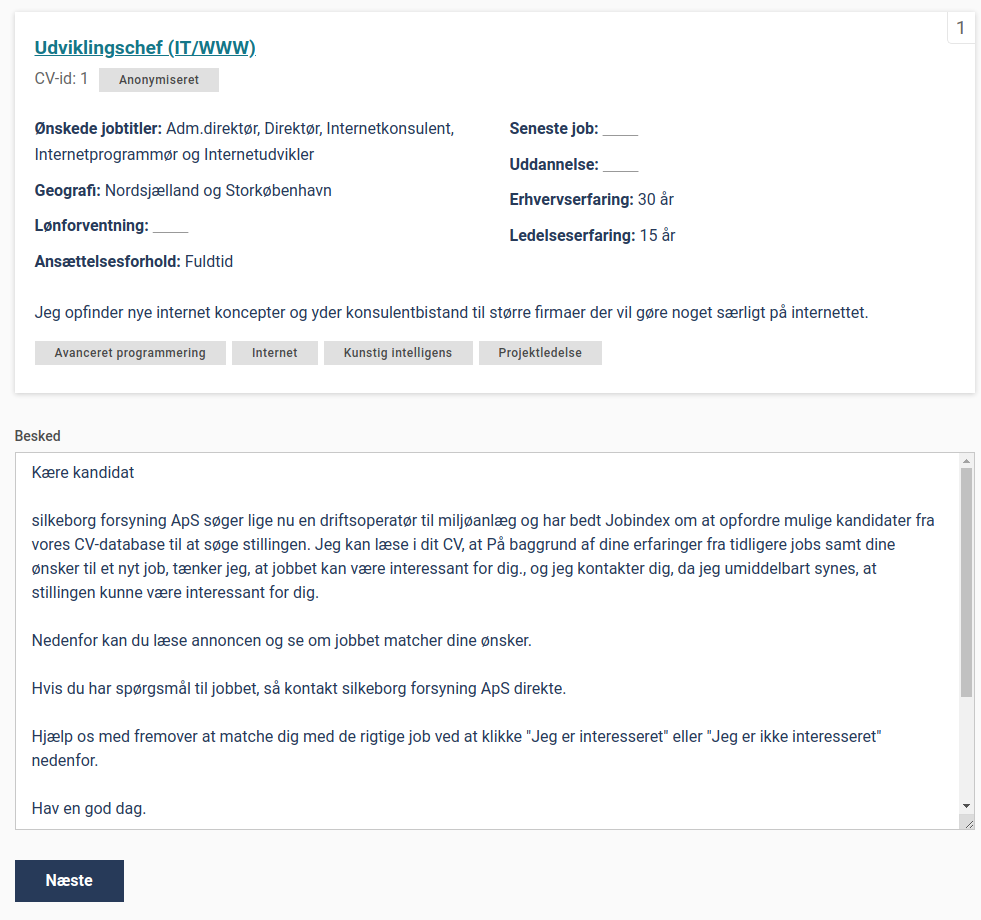}
         \caption{Helped case. }
         \label{fig:helped-after}
     \end{subfigure}
        \caption{Screenshots of the interface in (a) ``unhelped'' and (b) ``helped'' case.}
        \label{fig:screenshots}
        
\end{figure*}

We conduct an offline evaluation with real expert recruiters to evaluate the performance of the template-based email generation algorithm. The quantitative measures for natural language generation are not employed, since the template-based approach may produce significantly different structures from the real recruitment emails and still be reasonably good. Indeed, it remains an open question to design quantitative measures for evaluating the generated recruitment emails, as the judgments on the persuasiveness or personalization of the email vary from person to person.


We randomly sample a collection of truly matched (job, candidate) pairs from the Jobindex database. They are randomly assigned to each of a group of 10 recruiters. The pairs are randomly split into 5 sessions of 10 pairs each. Each recruiter is randomly assigned with ``helped'' tasks (with generated texts) and ``unhelped''  tasks (with generated templates).  Each task is ``helped'' for half of the recruiters and ``unhelped'' for the other half.  The interface contained pairs of matched job postings and candidate profiles, and an interface is provided to the recruiter with either pre-generated template (``unhelped'') or the email (``helped''). For the email, the inserted case-specific information is wrapped by brackets ``\{\}''. A checkbox is also presented to the recruiters for providing a 4-level judgment on the quality of the email or template (4 = Perfect, 3 = Minor Revision, 2 = Major Revision, 1 = Useless). The judgment is on language quality for templates and on both language quality and information correctness for emails. The recruiters are asked to click a button before and after writing the recruitment email, so that the difference between the recorded time stamps are the time spent on it. The screenshots of the interface are shown in Fig.~\ref{fig:screenshots}.

\begin{figure}[h]
\caption{Bar plots of annotations for the offline-evaluation. The red bars are the recruiters' judgments on the quality of generated templates, and blue bars are the recruiters on the quality of generated messages.}
\label{fig:eval_annotation_apr}
\centering
\includegraphics[width=0.47\textwidth]{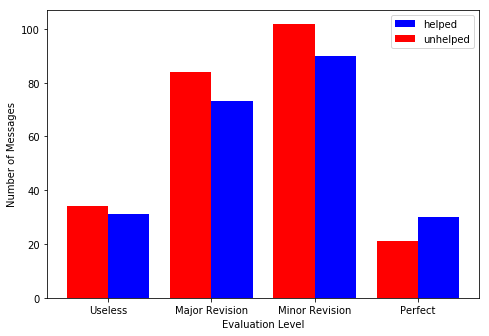}
\end{figure}

\noindent \textbf{Result on text quality}. We have collected the results of 224 helped tasks (with generated emails by our algorithm) and 241 unhelped tasks (with our generated templates by our algorithm). The recruiters' annotations on the quality of the generated texts are shown in Fig.~\ref{fig:eval_annotation_apr}. In terms of average ratings, we obtained a value of 2.53 for generated emails and 2.46 for generated templates for the evaluation of the new template-based system. Over 50\% of the generated texts (53.58\% for emails and 51.03\% for templates) are satisfactory, requiring minor or no edits. It is interesting to see that more recruiters are satisfied with the generated texts than the templates. This implies that the generated motivational sentence is exceptionally helpful for recruiters.  In the same time, there is still room for the text quality to improve, and neural method for generating case-specific information is a feasible direction.

\noindent \textbf{Result on time cost}. We computed the average time a recruiter spent on writing the message in the unhelped and helped case, respectively. The time difference between the two scenarios is the reduction of recruiters' time by our template-based automatic text generation system. As a consequence, recruiters spent an average of 178.99s on each matched pair in the helped case, and an average of 235.63s on each matched pair in the unhelped case. On average, a recruiter spends 56.64s shorter on writing an email with automatically generated messages than merely on top of the provided templates. In order to remove the discrepancy across individual recruiters, we zero-averaged the time for each recruiter and recalculated the time difference.  A closer but still remarkable gap between the times (51.31s) is observed. It shows that this simple template-based system can significantly save recruiters time on writing recruitment emails.

\section{Conclusion}

We have proposed recruitment email generation for job recruitment, a novel task in text generation. We demonstrate its challenge and significance in the pilot study and a user study of our template-based approach. The challenge mainly resides in constructing case-specific components of the email from the input job and candidate to enhance personalization and persuasiveness. Quite often, the reasons of a good match are semantically non-explicit and cannot be extracted by common word or phrase matching. At the same time, we have observed that a simple approach can remarkably benefit recruiters by saving around 1/4of their time.

Future work could contribute to this task in the following aspects:

\begin{itemize}
    \item{End-to-end neural generation of motivational sentences. Rather than the simple rule-based algorithm for composing the motivational sentences, we may learn a neural generation system from the real emails in a data-driven manner. }
    
    \item{Robust intrinsic evaluation metrics. The n-gram matching-based metrics, such as BLEU or ROUGE, are not suitable for this task. Beyond the real-user evaluation, it remains an open challenge to propose robust metrics that objectively evaluate the generated text, in terms of both language quality and task-related aspects.}
    
    \item{Deep representations of job and candidate to better extract reasons of match. Deep neural models should be constructed to perform deep understanding of jobs and candidates in order to support extraction of below-the-surface matched qualifications.}
    
    \item {Generating explanation of recommendations. Currently, the email generation is a separate process from job recommendation. It would be interesting to view recruitment emails instead as explanations of the recommendation systems, and propose recommendation models that supports interpretations of the recommendations. The authentic explanations of recommendation will be then be incorporated in the recruitment email.}

\end{itemize}

\section*{Acknowledgements}
This research was supported by the Innovation Fund Denmark, grant no. 0175-000005B.

\bibliography{anthology,custom}
\bibliographystyle{acl_natbib}

\appendix



\end{document}